%% file: bare_jrnl.tex
\documentclass[journal]{IEEEtran}

\usepackage{amsmath,graphicx}
\usepackage[utf8]{inputenc}
\usepackage[greek,english]{babel}
\usepackage[LGR, T1]{fontenc}
\usepackage{tikz}
\graphicspath{{gfx/}} 
\newcommand{\bs}{\mathbf}
\usepackage{amsmath,amssymb}
\DeclareMathOperator{\E}{\mathbb{E}}
\usepackage[bookmarks=false]{hyperref}
\usepackage{booktabs}
\usepackage{multirow}
\usepackage{soul}

\usepackage{subcaption}
\usepackage[]{graphicx} 

\input{mihalisan.tex}

\begin{document}
\title{End-to-End Multimodal Emotion Recognition  \\ using Deep Neural Networks}

\author{Panagiotis Tzirakis, 
        George Trigeorgis, 
        Mihalis A. Nicolaou, 
        Bj\"orn Schuller, 
        and~Stefanos~Zafeiriou} 

\markboth{Journal of \LaTeX\ Class Files,~Vol.~14, No.~8, August~2015}%
{Shell \MakeLowercase{\textit{et al.}}: Bare Demo of IEEEtran.cls for IEEE Journals}

\maketitle


\begin{abstract}
Automatic affect recognition is a challenging task due to the various modalities emotions can be expressed with. Applications can be found in many domains including multimedia retrieval and human computer interaction. In recent years, deep neural networks have been used with great success in determining emotional states. Inspired by this success, we propose an emotion recognition system using auditory and visual modalities. To capture the emotional content for various styles of speaking, robust features need to be extracted. To this purpose, we utilize a Convolutional Neural Network (CNN) to extract features from the speech, while for the visual modality a deep residual network (ResNet) of 50 layers. In addition to the importance of feature extraction, a machine learning algorithm needs also to be insensitive to outliers while being able to model the context. To tackle this problem, Long Short-Term Memory (LSTM) networks are utilized. The system is then trained in an end-to-end fashion where – by also taking advantage of the correlations of the each of the streams – we manage to significantly outperform the traditional approaches based on auditory and visual handcrafted features for the prediction of spontaneous and natural emotions on the RECOLA database of the AVEC 2016 research challenge on emotion recognition.
\end{abstract}

\begin{IEEEkeywords}
end-to-end learning, emotion recognition, deep learning
\end{IEEEkeywords}

\IEEEpeerreviewmaketitle

\section{Introduction}
\IEEEPARstart{E}{motion} recognition is an essential component towards complete interaction between human and machine, as affective information is fundamental to human communication. Applications of emotion recognition can be found in different domains. For instance, emotion states can be used to monitor and predict fatigue state~\cite{ji2004real}. In speech recognition, emotion recognition can be used in call centres, where the goal is to detect the emotional state of the caller and provide feedback for the quality of the service~\cite{burkhardt2006detecting}.

The task of recognising emotions is challenging because human emotions lack of temporal boundaries and different individuals express emotions in different ways~\cite{anagnostopoulos2015features}. Although current work around emotion recognition was concentrated around infering the emotion of a subject out of its speech other modalities such as visual information~(facial gestures) have also been used.
With the advent of deep neural networks in the last decade a number of groundbreaking improvements have been observed in several established pattern recognition areas such as object, speech and speaker recognition, as well as in combined problem solving approaches, e.g. in audio-visual recognition, and in the rather recent field of paralinguistics. 

Numerous studies have shown the favourable property of these network variants to model inherent structure contained in the speech signal~\cite{hinton2012deep}, with
more recent research attempting \textit{end-to-end} optimisation utilising as little human a-priori knowledge as possible~\cite{graves2014towards}. Nevertheless, the majority of these works make use of commonly hand-engineered features have been used as input features, such as Mel-Frequency Cepstral Coefficients (MFCC), Perceptual Linear Prediction (PLP) coefficients, and supra-segmental features such as those used in the series of ComParE~\cite{Schuller13-TI2} and AVEC challenges~\cite{avec2015}, 
which build upon knowledge gained in decades of auditory research and have shown to be robust for many speech domains.

Recently, however, a trend in the machine learning community has emerged towards deriving a representation of the input signal directly from \textit{raw}, unprocessed data. 
The motivation behind this idea is that, ultimately, the network learns an intermediate representation of the raw input signal automatically that better suits the task at hand and hence leads to improved performance. 

In this paper, we study automatic affect sensing using both speech and visual information in an end-to-end manner. Features are extracted from the speech signal using a CNN architecture designed for the audio channel and from the visual information using a ResNet-50 network architecture~\cite{he2016deep}. The output of these networks are fused together and fed to an LSTM to find the affective state of individuals. Contrary to the current practises, where each network is trained individually and the results are simply fed to a classifier our system is trained in an end-to-end manner. To our knowledge this is the first work in literature that applies such an end-to-end model for audio visual emotion recognition task. Furthermore, we suggest using explicit maximisation of the {\it concordance correlation coefficient} ($\rho_c$)~\cite{Lin89-ACC} in our model and show that this improves performance in terms of emotion prediction compared to optimising the mean square error objective, which is traditionally used. Finally, by further studying the activations of different cells in the recurrent layers, we find the existence of interpretable cells, which are highly correlated with several prosodic and acoustic features that were always assumed to convey affective information in speech, such as the loudness and the fundamental frequency. A preliminary version of this work was presented in \cite{trigeorgis2016adieu}, where only the raw speech waveform was used. We extend this work by considering also the visual modality in an end-to-end manner.

To show the benefit of our proposed multimodal model we evaluated it in the REmote COLlaborative and  Affective (RECOLA) database. A part of this database was used for the Audio/Visual Emotion Challenge and Workshop (AVEC) 2016. Our model is trained using the whole database. Results show that the multimodal model benefits from the two modalities by producing equal results for arousal and valence as the speech and visual networks, respectively. We compare the unimodal and the multimodal models using results obtained in the AVEC 2016 challenge. Only the papers that used the audio, visual or audiovisual modalities are considered. In order to perform a fair comparison we apply the proposed method on the test set of the AVEC challenge. As shown by our experiments our unimodal models produce the best results for both the speech and visual modalities. 

The remainder of the paper is organized as follows. Section II reports related studies on emotion recognition using multiple modalities with DNNs. Section III introduces the multimodal model architecture. Section IV describes the dataset used for the experiments. Section V presents the experiments performed and reports the results. Finally, section VI concludes this paper. 

\section{Related Work}

The performance of pattern recognition models have been greatly improved with DNNs. Recently, a series of new neural network architectures have been revitalised, such as autoencoder networks~\cite{zemel1994autoencoders}, Convolutional Neural Networks (CNNs)~\cite{lecun1989generalization}, Deep Belief Networks (DBNs)~\cite{hinton2006fast} or memory enhanced neural network models such as Long Short-Term Memory (LSTM)~\cite{hochreiter1997long} models. 

These models have been used in various ways for multimodal recognition tasks such as in speech recognition. For instance, Ngiam et al. \cite{ngiam2011multimodal} proposed a Multimodal Deep Autoencoder (MDAE) network to extract features from audio and video modalities. First, a bimodal DBN was trained to initialize the deep autoencoder and then the MDAE was fine-tuned to minimize the reconstruction error of both modalities. In another study, Hu et al.~\cite{hu2016temporal} proposed a temporal multimodal network named Recurrent Temporal Multimodal Restricted Boltzmann Machine (RTMRBM) to model audiovisual sequence of data. Another task that DNNs have also been used is gesture recognition. In \cite{wu2016deep} the authors use skeletal information and RGB-D images to recognize gestures. More particularly, they use DBNs to process skeleton features and a 3D CNN for the RGB-D data. Temporal information is considered by stacking a Hidden Markov Model (HMM) on top.

The emotion recognition domain has highly benefited with the advent of DNNs. Some works explore deep learning approaches for speech emotion recognition. For instance, Han et al. \cite{han2014speech} uses hand-crafted features to feed a DNN that produces a probability distribution over categorical emotion states. From these probabilities they compute statistics from the whole utterance and finally, they perform classification by training an extreme learning machine. Lim et al \cite{lim2016speech} after transforming the data using short time fourier transform, they used CNNs to extract high-level features. In order to capture the temporal structure LSTMs were used. In a similar work, Trigeorgis et al.\cite{trigeorgis2016adieu} proposed an end-to-end model that uses a CNN to extract features from the raw signal and then an LSTM network to capture the contextual information in the data. 

Other works try to solve the emotion recognition task by using facial information with DNNs. For example, Huang et al. \cite{huang2016deep} proposed a transductive learning framework for image-based emotion recognition by combining DNNs and hypergraphs. More particularly, after the DNN was trained for emotion classification task, each node in the last fully connected layer was considered as an attribute and used to form a hyperedge in a hypergraph. In another study, Ebrahimi et al. \cite{ebrahimi2015recurrent} combined CNNs and RNNs to recognise categorical emotions in videos. A CNN was first trained to classify static images containing emotion. Then, the extracted features from the CNN were used to train an RNN to produce an emotion for the whole video.


Recently, combining audio and visual modalities have great success for recognizing emotions. Some studies exploited the beneficial features DNNs can extract \cite{liu2016emotion,kim2013deep,kahou2016emonets}. Kim et al. \cite{kim2013deep} proposed four different DBN architectures with one of them being a basic 2-layer DBN, and the others variation of it. The basic architecture first learns the features of the audio and video separately. After which it concatenates these features from the two modalities it uses them to learn the second layer. The features were evaluated using a Support Vector Machine (SVM). In another study, Kahou et al. \cite{kahou2016emonets} proposed to combine modality-specific DNNs to recognize categorical emotions in video. A CNN was used to analyze the video frames, a DBN to capture audio information, a deep autoencoder  to  model  human  actions  depicted  within  the  entire  scene, and finally a CNN network to extract features from the mouth of the human. To output a final prediction they used two techniques that gave similar results. The first is to take the average of modality-specific predictions and in the second they learned an SVM with an RBF kernel using the concatenation features. In another study \cite{sun2016exploring} compared hand-crafted features extracted from faces using multi-scale Dense SIFT features (MSDF), and features extracted from CNNs to train linear Support Vector Regression (SVR). The extracted audio features were the extended Geneva Minimalistic Acoustic Parameter Set (eGeMAPS). The combination of the features were used to learn a Support Vector Regression (SVR).

Zhang et al.~\cite{Zhang} used a multimodal CNN for classifying emotions with audio and visual modalities. The model is trained in two phases. In the first phase the two CNNs are pretrained on large image datasets and fine-tuned to perform emotion recognition. The audio CNN takes as input the mel-spectogram segment of the audio signal and the video CNN takes the face. In the second phase a DNN was trained that comprised of a number of fully-connected layers. The concatenation of the features extracted by the two CNNs were inputted. In another study, Ringeval et al.~\cite{ringeval2015prediction} uses an BLSTM-RNN to capture the contextual information that exists in the multimodal features (audio, video, physiological) extracted from the data. In a more recent work, Han et al.~\cite{han2016} proposes the strength modeling framework which can be implemented as feature-level and decision-level fusion strategy and comprises of two regression models. The first model’s predictions are concatenated with the original feature vector and fed to the second regression model for the final prediction. 

The importance of recognizing emotions motivated the creation of the Audio/Visual Emotion Challenge and Workshop (AVEC)~\cite{valstar2016avec}. In 2016 challenge audio, video and physiological modalities were considered. In one of the submitted models for the challenge, Huang et al. \cite{huang2016staircase} proposed to use variants of Relevance Vector Machine (RVM) for modeling audio, video and audiovisual data. In another work model by Weber et al.~\cite{Weber2016} used high-level geometry features for predicting dimensional features. Brady et al.~\cite{brady2016multi} also used low- and high-level features for modeling emotions. In a different study Povolny et al.~\cite{povolny2016multimodal} complemented original baseline features for both audio and video to perform emotion recognition. Somandepalli et al.~\cite{somandepalli2016online} also used additional features but only for the audio modality.

All of the works in the literature make use of commonly hand-crafted features in audio or visual modality or in some cases in both of them. Moreover, they do not always consider temporal information in the data. In this study we propose a multimodal model trained end-to-end that also considers the contextual temporal information.

\section{Proposed Method}

One of the first steps in a traditional machine learning algorithms is to extract features from the data. To extract features in audio, finite impulse response filters can be used which perform time-frequency decomposition to reduce the influence of background noise~\cite{hirsch1991improved}. More complicated hand-engineered kernels, such as gammatone filters~\cite{schluter2007gammatone}, which were formulated by studying the frequency responses of the receptive fields of auditory neurons of grassfrogs, can be used as well. 

A key component of our model is the convolution operation. For the audio and visual signals, 1-d and 2-d convolution is used, respectively.
\begin{equation}
(f \star h)(i,j) = \sum_{k=-T}^T \sum_{m=-T}^T f(k,m) \cdot h(i - k, j - m)
\end{equation}
where $f(x)$ is a kernel function whose parameters are learnt from the data of the task in hand. 
After the spatial-modeling of the signals, which removes background noise and enhances specific parts of the signals for the task in hand, we model the temporal structure of both speech and video by using a recurrent network with LSTM cells. We use LSTM for \textit{(i)} simplicity, and \textit{(ii)} to fairly compare against existing approaches which concentrated in the combination of hand-engineered features and LSTM networks. Finally, our model is subsequently trained with backpropagation using the objective function, cf.~\autoref{eq:ccc_loss}.

\subsection{Visual Network}

None of the first steps in the traditional face recognition pipeline is feature extraction utilizing hand-crafted representations such as Scale Invariant Feature Transform (SIFT) and Histogram of Oriented Gradients (HOG). Recently, deep convolutional networks have been used to extract features from faces~\cite{Zhang}.


In this study we use a deep residual network (ResNet) of 50 layers~\cite{he2016deep}. As input to the network we used the pixel intensities from the cropped faces of the subject's video. Deep residual networks adopt residual learning by stacking building blocks of the form:

\begin{equation}
\bs y_k = \mathcal{F}( \bs x_k,\{W_k\}) + \bs h(x_k)
\end{equation}

where $\bs x$ and $\bs y$ are the input and output of the layer $k$, $\mathcal{F}( \bs x_k,\{W_k\})$ is the residual function to be learned and $h(\bs x_k)$ can be either an identity mapping or a linear projection to match the dimensions of function $\mathcal{F}$ and the input $\bs x$.

The first layer of ResNet-50 is a $7x7$ convolutional layer with 64 feature maps, followed by a max pooling layer of size $3x3$. The rest of the network comprises of 4 bottleneck architectures, where after these architectures a shorcut connection is added. These architectures contain 3 convolutional layers of sizes 1x1, 3x3, and 1x1, for each residual function. \autoref{bottleneck_architectures} shows the replication and the sizes of the feature maps for each bottleneck architecture. After the last bottleneck architecture an average pooling layer is inserted.

\begin{table}[ht]
\centering
\begin{tabular}{l|l|l}\toprule\midrule
 Bottleneck layer & Replication & \parbox[t]{3cm}{Number of feature maps\\ (1x1, 3x3, 1x1) } \\\midrule
 1st & 3 & 64, 64, 256 \\
 2nd & 4 & 128, 128, 512 \\
 3rd & 6 & 256, 256, 1024 \\ 
 4th & 3 & 512, 512, 2048 \\ 
\end{tabular}
\caption{The replication of each bottleneck architecture of the ResNet-50 along with the size of the features maps of the convolutions.}
\label{bottleneck_architectures}
\end{table}



\subsection{Speech Network}
In contrast to previous work done in the field of paralinguistics, 
where acoustic features are first extracted and then passed to a machine learning algorithm, 
we aim at learning the feature extraction and regression steps in one jointly trained model for predicting the emotion. 

\textit{Input}. We segment the raw waveform to 6\,s long sequences after we preprocess the time-sequences to have zero mean and unit variance to account for variations in different levels of loudness between the speakers. At 16\,kHz sampling rate, this corresponds to a 96000-dimensional input vector. 

\textit{Temporal Convolution}. We use $F=20$ space time finite impulse filters with a 5ms window in order to extract fine-scale spectral information from the high sampling rate signal.

\textit{Pooling across time}. The impulse response of each filter is passed through a half-wave rectifier (analogous to the cochlear transduction step in the human ear) and then downsampled to 8\,kHz by pooling each impulse response with a pool size = 2. 

\textit{Temporal Convolution}. We use $M=40$ space time finite impulse filters of 500ms window. These are used to extract more long-term characteristics of the speech and the roughness of the speech signal. 

\textit{Max pooling across channels}. We perform max-pooling across the channel domain with a pool size of 10. This reduces the dimensionality of the signal while preserving the necessary statistics of the convolved signal.

\textit{Dropout}. Due to the large number of parameters (1.1 mil.) compared to the number of training examples we need to perform some regularisation in order for the model not to overfit on the training data. We opt to use dropout with a probability of 0.5. 

\subsection{Objective function}
To evaluate the agreement level between the predictions of the network and the gold-standard derived from the annotations, 
the concordance correlation coefficient ($\rho_c$)~\cite{Lin89-ACC} has recently been proposed~\cite{Ringeval15-POA,avec2015}. 
Nonetheless, previous work minimized the MSE during the training of the networks, but evaluated the models with respect to $\rho_c$~\cite{Ringeval15-POA,avec2015}. 
Instead, we propose to include the metric used to evaluate the performance in the objective function ($\mathcal{L}_{c}$) used to train the networks.  
Since the objective function is a cost function, we define $\mathcal{L}_{c}$ as follow:

\begin{align}\nonumber
\mathcal{L}_{c}
&= 1 - \rho_c = 1 - \dfrac{2 \sigma_{xy}^2}{\sigma_x^2 + \sigma_y^2 + (\mu_x - \mu_y)^2} \\
&= 1 - 2 \sigma_{xy}^2 \psi^{-1} 
\label{eq:ccc_loss}
\end{align}
where $\psi = \sigma_x^2 + \sigma_y^2 + (\mu_{x} - \mu_{y})^2$ and $\mu_{x} = \E(\bs x)$, ${\mu_{y} = \E(\bs y)}, {\sigma_x^2 = \mbox{var}(\bs x)}, \sigma_y^2 = \mbox{var}(y)$ and $\sigma^2_{xy} = \mbox{cov}(\bs x, \bs y)$
Thus, to minimise $\mathcal{L}_{c}$ (or maximise $\rho_c$), 
we backpropagate the gradient of the last layer weights with respect to $\mathcal{L}_{c}$,

\begin{equation}
\dfrac{\partial \mathcal{L}_{c}}{\partial \bs x} \propto 2\dfrac{\sigma_{xy}^2 (\bs x - \mu_y)}{\bs \psi^2} + \dfrac{\mu_y - \bs y}{\psi},
\end{equation}

where all vector operations are done element-wise.

\subsection{Network Training}

Before training the multimodal network, each modality-specific network is trained separately to speed up the training procedure.


\textit{Visual Network}. For the visual network, we chose to fine-tune the pretrained ResNet-50 on the database used in this work. This model was trained on the ImageNet 2012 \cite{russakovsky2015imagenet} classification dataset that consists of 1000 classes. The pretrained model was preferred than training the network from scratch in order to be benefited by the features already learned by the model. To train the network a 2-layer LSTM, with 256 cells each is stack on top of it to capture temporal information.

\textit{Speech Network}. The CNN network operates on raw signal to extract features from. In order to consider the temporal structure of speech, we use two LSTM layers with 256 cells each on top of the CNN.

\begin{figure*}[h]
\centering
\includegraphics[width=16cm, height=9cm]{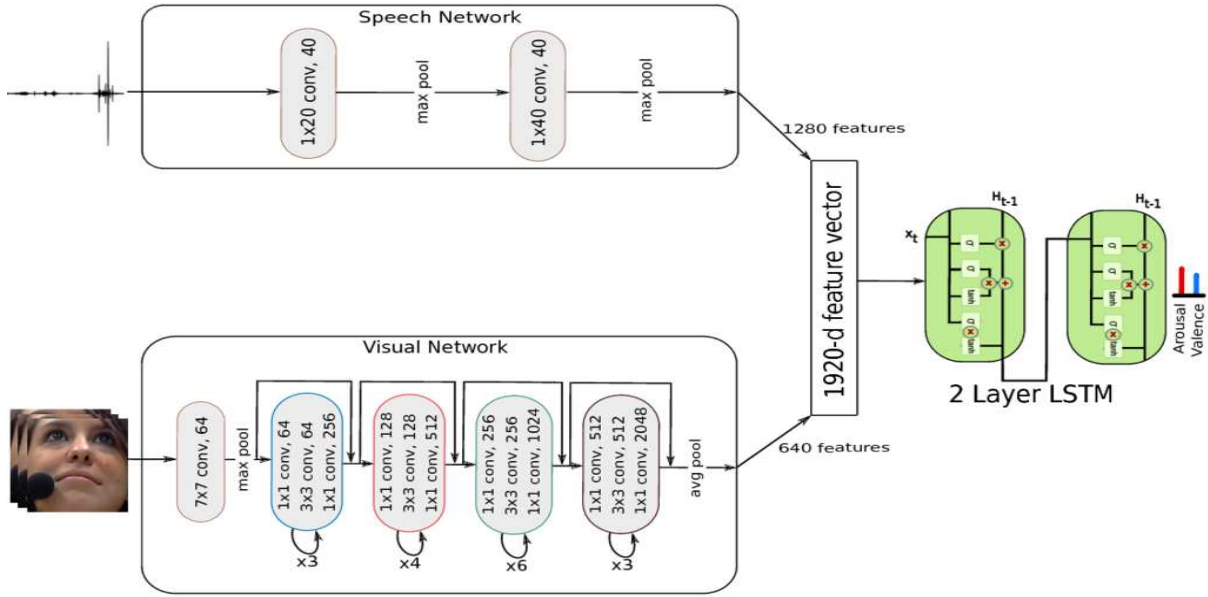}
\caption{The network comprises of two parts: the multimodal feature extraction part and the RNN part. The multimodal part extracts features from raw speech and visual signals. The extracted features are concatenated and used to feed 2  LSTM layers. These are used to capture the contextual information in the data.}
\label{fig:multi_network}
\end{figure*}

\textit{Multimodal Network}. After training the visual and speech networks the LSTM layers are discarded and only the extracted features are considered. The speech network extracts 1280 features while the visual netowork 640 features. These are concatenated to form a 1920 dimensional feature vector and used to feed a 2-layer LSTM with 256 cells each. The LSTM layers are trained and the visual and speech networks are fine-tuned. \autoref{fig:multi_network} shows the multimodal network.

The goal for each unimodal and the multimodal network is to minimize:

\begin{equation}
\mathcal{L}_{c} = \frac{\mathcal{L}_{c}^{a} + \mathcal{L}_{c}^{v}}{2}
\end{equation}

where $\mathcal{L}_{c}^{a}$ and $\mathcal{L}_{c}^{v}$ are the concordance of the arousal and valence, respectively. 

For the recurrent layers of speech, visual and multimodal networks, we segment the 6\,s sequences to 150 smaller sub-sequences to match the granularity of the annotation frequency of 40\,ms.

\section{Dataset}

Time-continuous prediction of spontaneous and natural emotions (arousal and valence) is investigated on speech and visual data by using the REmote COLlaborative and  Affective (RECOLA) database introduced by Ringeval  et  al. ~\cite{Ringeval13-ITR}; the full dataset for  which  participants  gave their consent to share their data is used for the purpose of this study. Four modalities are included in the corpus: audio,  video,  electro-cardiogram  (ECG)  and  electro-dermal  activity  (EDA). In total, 9.5h of multimodal recordings from 46 French-speaking participants were recorded and annotated for 5\,minutes each, performing a collaboration task in dyads during a video conference. Among the participants, 17  were  French,  three  German  and three  Italian. The dataset is split in three partitions -- train (16 subjects), validation (15 subjects) and test (15 subjects) -- by stratifying (i.e., balancing) the gender and the age of the speakers. Finally, 6 French -speaking annotators (three male, three female) annotated all the recordings.

\section{Experiments \& Results}

For training the models we utilised the Adam optimisation method~\cite{kingma2014adam}, and a fixed learning rate of $10^{-4}$ throughout all experiments. For the audio model we used a mini-batch of 25 samples. 
Also, for regularisation of the network, we used dropout~\cite{srivastava2014dropout} with $p=0.5$ for all layers except the recurrent ones. 
This step is important as our models have a large amount of parameters ($\approx{1.5M}$) and not regularising the network makes it prone on overfitting on the training data. 

For the video model, the image size used was $96 \times 96$ with mini-batch of size 2. Small mini-batch is selected because of hardware limitations. The data were augmented by resizing the image to size $110 \times 110$ and randomly cropping it to equal its original size. This produces a scale invariant model. In addition, color augmentation is used by introducing random brightness and saturation to the image. 

Finally, for all investigated methods, a chain of post-processing is applied to the predictions obtained on the development set: \textit{(i)} median filtering (with size of window ranging from 0.4\,s to 20\,s)~\cite{avec2015}, \textit{(ii)} centring (by computing the bias between gold-standard and prediction)~\cite{Kachele15-EMF}, \textit{(iii)} scaling (using the ratio of standard-deviation of gold-standard and prediction as scaling factor)~\cite{Kachele15-EMF} and \textit{(iv)} time-shifting (by shifting the prediction forward in time with values ranging from 0.04\,s to 10\,s), to compensate for delays in the ratings~\cite{Mariooryad14-CTC}. Any of these post-processing steps is kept when an improvement is observed on the $\rho_c$ of the validation set, and applied then with the same configuration on the test partition.

\subsection{Ablation study}
Due to memory and training instability concerns~\cite{cho2014learning} its not always optimal to use very large sequences in recurrent networks. The justification for this can be either the over-blowing of gradients or the very deep unrolled graph which makes training of such big networks harder.

In order to choose the best sequence length to feed our LSTM layers, we conducted experiments using  sequence lengths 75, 150, and 300 for both speech and visual models.~\autoref{tab_seqLength} shows the results on the development set. For all the experiments a total of 60 epochs were run for our models.

For the visual network we expect to get the highest value in the valence dimension, while for the speech model in the arousal dimension. Results indicate that the best value for the speech model is 150 while for the visual model is 300. Due to the fact that the difference in performance for the visual network is small when sequence length of 150 or 300 is used, we chose to train the mutlimodal network with 150 sequence length. 



\begin{table}[ht]
\centering
\begin{tabular}{l|l|l}\toprule\midrule
 \parbox[t]{1cm}{Sequence\\ length}  &  Arousal & Valence\\\midrule
 \multicolumn{3}{l}{\em\small Visual network} \\\midrule
 75  & .293 & .276 \\
 150 & .363 & .488 \\
 300 & .193 & .496 \\ \midrule
 \multicolumn{3}{l}{\em\small  Speech network} \\\midrule
 75  & .727 & .345 \\
 150 & .744 & .369 \\
 300 & .685 & .130 \\
\end{tabular}
\caption{Results (in terms of $\rho_c$) on arousal and valence after 60 epochs when varying sequence length for speech and visual networks.}
\label{tab_seqLength}
\end{table}

\subsection{Speech Modality}
Results obtained for each method, using all 46 participants, are shown in~\autoref{tab_audio}. 
In all of the experiments, our model outperforms the designed features in terms of $\rho_c$.
One may note, however, that the eGEMAPS feature set provides close performance on valence, which is much more difficult to predict from speech compared to arousal.
Furthermore, we show that by incorporating $\rho_c$ directly in the optimisation function of all networks allows us to optimise the models on the metric $(\rho_c)$ on which we evaluate the models. This provides us with \textit{i)} a more elegant way to optimise models, and \textit{ii)} gives consistently better results across all test-runs as seen in~\autoref{tab_audio}.

\begin{table}[ht]
\centering
\begin{tabular}{l|l|l|l}\toprule\midrule
 Predictor & Features   & Arousal & Valence\\\midrule
 \multicolumn{4}{l}{\em\small a. Mean squared error objective} \\\midrule
 SVR   & eGeMAPS  & .318~(.489) & .169~(.210)\\
 SVR   & ComParE  & .366~(.491) & .180~(.178)\\
 BLSTM & eGeMAPS  & .300~(.404) & .192~(.187)  \\
 BLSTM & ComParE  & .132~(.221) & .117~(.152) \\
 Proposed  & raw signal & .684~({.728}) &  {.249}~({.312})  \\\toprule\midrule
\multicolumn{4}{l}{\em\small b. Concordance correlation coefficient objective} \\\midrule
 BLSTM & eGeMAPS & .316~(.445) & .195~(.190)  \\
 BLSTM & ComParE & .382~(.478) & .187~(.246) \\
 \midrule
 Proposed  & raw signal & \textbf{.699}~(\textbf{.752}) &  \textbf{.311}~(\textbf{.406})  \\
\end{tabular}
\caption{RECOLA dataset results (in terms of $\rho_c$) for prediction of arousal and valence. In parenthesis are the performance obtained on the development set. In \textit{a)} we optimised the models wrt. MSE whereas in \textit{b)} wrt. $\rho_c$.}
\label{tab_audio}
\end{table}

In addition, we compare the performance on the results obtained for methods that exist in the literature. Most of them have been submitted to the AVEC 2016 challenge, using 27 participants~\autoref{tab_audio_avec}. In case performance on the test or validation set was not reported in the paper a dash is inserted. Results show that our model outperforms the other models in the test set when predicting arousal dimension. It is important to notice that although our model gets lower $\rho_c$ on the arousal dimension for the validation set compared to the baseline of the challenge, its performance is better on the test set.

\begin{table}[ht]
\centering
\begin{tabular}{l|l|l|l}\toprule\midrule
 Predictor & Features   & Arousal & Valence\\\midrule
 Baseline~\cite{valstar2016avec} & eGeMAPS & .648~(.796) & \textbf{.375}~(.455) \\
 RVM~\cite{huang2016staircase} & eGeMAPS & -~(.750) & -~(.396) \\
 Audio BN-Multi~\cite{povolny2016multimodal} & Mixed & -~(.833) & -~(.503) \\ 
 Brady et al~\cite{brady2016multi} & MFCC & -~({.846}) & -~(.450) \\ 
 Weber et al.~\cite{Weber2016} & eGeMAPS & -~(.793) & -~(.456) \\
 Somandepalli et al.~\cite{somandepalli2016online} & Mixed & -~(.800) & -~(.448) \\ 
 Han et al.~\cite{han2016} & 13 LLDs & .666~(.755) & .364~({.476}) \\
 \midrule
 Proposed  & raw signal & \textbf{.715}~({.786}) &  {.369}~(.428)  \\
\end{tabular}
\caption{RECOLA dataset results (in terms of $\rho_c$) for prediction of arousal and valence. In parenthesis are the performance obtained on the development set. A dash is inserted if the results were not reported in the original papers.}
\label{tab_audio_avec}
\end{table}

\subsubsection{Relation to existing acoustic and prosodic features}
\label{sec:gate_activations}

The speech signals convey information about the affective state either explicitly, i.e., by linguistic means, or implicitly, i.e., by acoustic or prosodic cues. 
It is well accepted amongst the research community that certain acoustic and prosodic features play an important role in recognising the affective state~\cite{Scherer03-VCO}. 
Some of these features, such as the mean of the fundamental frequency (F0), mean speech intensity, loudness, as well as pitch range~\cite{Eyben15-TGM}, should thus be captured by our model.
\begin{figure}[h]
\centering
\includegraphics[width=\linewidth]{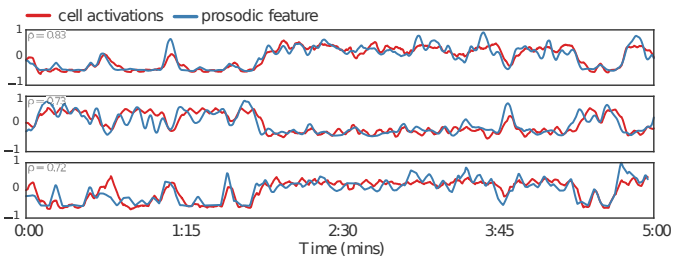}
\caption{A visualisation of three different gate activations vs. different acoustic and prosodic features that are known to affect arousal for an unseen recording to the network. From top to bottom:
range of RMS energy~($\rho=0.83$),
loudness~($\rho=0.73$),
mean of fundamental frequency~($\rho=0.72$)
}
\label{fig:gate_activations}
\end{figure}

To gain a better understanding of what our speech model learns, and how this relates to existing literature, we study the statistics of gate activations in the network applied on an unseen speech recording; a visualisation of the hidden-to-output connections of different cells in the recurrent layers of the network is given in~\autoref{fig:gate_activations}. This plot shows that certain cells of the model are very sensitive to different features conveyed in the original speech wave form. 

\subsection{Visual Modality}

Visual modality has been shown to more easily predict the valence dimension rather than the arousal. ~\autoref{tab_visual} presents the best results on the RECOLA dataset for the valence dimension. Only the work from Han et al.~\cite{han2016} was not submitted to the AVEC 2016 challenge. The features used for all of the models are appearance and geometric. For appearance Local Gabor Binary Patterns from Three Orthogonal Planes (LGBP-TOP) features were extracted, whereas facial landmarks were extracted for the geometric features. The input to our network are the raw pixel intensities from the face extracted from the frames of the videos using the Multi-Domain Convolutional Neural Network Tracker (MDNet)~\cite{nam2016mdnet} tracking algorithm. This algorithm takes the bounding box of the face in the first frame of the video and tracks it in all frames.

As expected visual modality benefits the models in the valence dimension. The only exception is the Video CNN-L4 model which performs better in the arousal dimension when appearance features are used. Our model outperforms all the other models in the valence dimension for the test set.

\begin{table}[ht]
\centering
\begin{tabular}{l|l|l|l}\toprule\midrule
 Predictor & Features   & Arousal & Valence\\\midrule
 Baseline~\cite{valstar2016avec} & Geometric & .272~(.379) & .507~({.612}) \\
 RVM~\cite{huang2016staircase} & Geometric & -~(.467) & -~(.571) \\
 Video CNN-L4~\cite{povolny2016multimodal} & Mixed & -~(.595) & -~(.497) \\ 
 Brady et al~\cite{brady2016multi} & Appearance & -~(.346) & -~(.511) \\
 Weber et al.~\cite{Weber2016} & Geometric & -~(.476) & -~(.683) \\
 Somandepalli et al.~\cite{somandepalli2016online} & Geometric & -~(.297) & -~(.612) \\ 
 Han et al.~\cite{han2016} & Geometric & .265~(.292) & .394~(.592) \\ 
 \midrule
Proposed  & raw signal & \textbf{.435}~(.371) &  \textbf{.620}~(.637)  \\
\end{tabular}
\caption{RECOLA dataset results (in terms of $\rho_c$) for prediction of arousal and valence. In parenthesis are the performance obtained on the development set. A dash is inserted if the results were not reported in the original papers.}
\label{tab_visual}
\end{table}

\subsection{Multimodal Analysis}

\begin{figure*}[th]

\begin{subfigure}[b]{1.05\textwidth}
\centering
   \includegraphics[width=12cm, height=4cm]{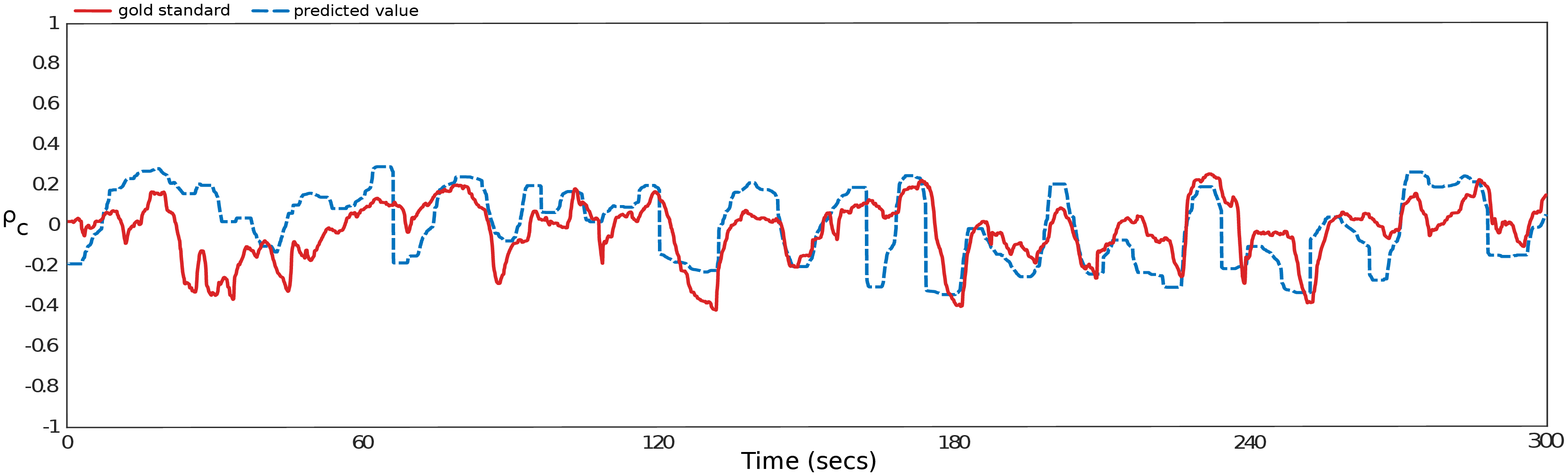}
   \caption{Arousal}
   \label{fig:arousal} 
\end{subfigure}

\begin{subfigure}[b]{1.05\textwidth}
\centering
   \includegraphics[width=12cm, height=4cm]{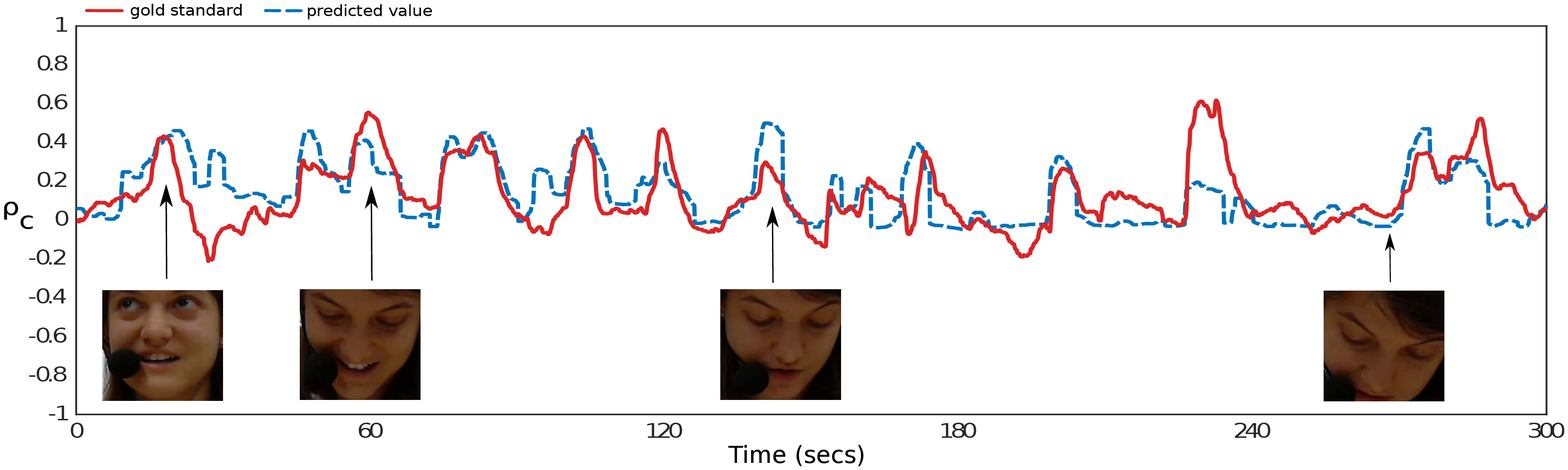}
   \caption{Valence}
   \label{fig:valence}
\end{subfigure}

\caption{The predicted and the gold standard for the arousal (a) and valence (b) on a video from the test set for 300 seconds.}
\label{multimodal}
\end{figure*}

Only two other models found in the literature to use both speech and visual modalities on the RECOLA database. These are the Output-Associative Relevance Vector Machine Staircase Regression (OA RVM-SR)~\cite{huang2016staircase} and the strength modeling system proposed by Han et al.~\cite{han2016}. Results are shown in~\autoref{tab_multi_results}. Our model outperforms the other two models in the valence dimension with high magnitude. For the arousal dimension the OA RVM-SR produces the best results. However, we would like to note here that there are two main differences between our system and the system in ~\cite{huang2016staircase} (a) the system in ~\cite{huang2016staircase} was trained using both training and validation set, whereas our model was trained using only the training set and (b) 
our system operates directly on the raw pixel domain, while the system in ~\cite{huang2016staircase} made use of a number of geometric features (e.g., 2D/3D facial landmarks etc.) which require the presence of an accurate facial landmark tracking methodology (ours was applied on the results of a conventional face detector only). We expect that our results would be improved further by applying similar strategies. Finally, to further demonstrate the benefits of our model for automatic prediction of arousal and valence~\autoref{multimodal} illustrates results for single test subject from RECOLA.

\begin{table}[ht]
\centering
\begin{tabular}{l|l|l|l|l}\toprule\midrule
 Predictor & \parbox[t]{0.5cm}{Audio\\Features}   & \parbox[t]{0.5cm}{Visual\\Features}  & Arousal & Valence\\\midrule
\parbox[t]{1cm}{OA \\ RVM-SR~\cite{huang2016staircase}}  &  \parbox[t]{0.5cm}{eGeMAPS\\ComParE} & \parbox[t]{0.5cm}{Geometric\\Appearance}  & \textbf{.770}~(.855) &  .545~(.642)  \\
\midrule
 Han et al.~\cite{han2016} & 13 LLDs & Geometric & .610~(.728) & .463~(.544) \\
 \toprule\midrule
Proposed  & raw signal & raw signal & .714~(.731) & \textbf{.612}~(.502)  \\
\end{tabular}
\caption{RECOLA dataset results (in terms of $\rho_c$) for prediction of arousal and valence. In parenthesis are the performance obtained on the development set.}
\label{tab_multi_results}
\end{table}

\section{Conclusion}
In this paper, we propose a multimodal system that operates on the raw signal, to perform an end-to-end spontaneous emotion prediction task from speech and visual data. To consider the contextual information in the data LSTM network was used. To speed up the training of the model we pretrained the speech and visual networks, separately. In addition, we study the gate activations of the recurrent layers in the speech modality and find cells that are highly correlated with prosodic features that were always assumed to cause arousal.
Our experiments on the unimodal modality show that our models achieve significantly better performance in the test set in comparison to other models using the RECOLA database including those submitted to the AVEC2016 challenge, thus demonstrating the efficacy of learning features that better suit the task-at-hand. On the other hand, our multimodal model performs better in the valence dimension rather than the arousal compared to other models.

\section*{Acknowledgment}

The support of the EPSRC Centre for Doctoral Training in High Performance Embedded and Distributed Systems (HiPEDS, Grant Reference EP/L016796/1) is gratefully acknowledged.

\ifCLASSOPTIONcaptionsoff
  \newpage
\fi



\bibliographystyle{IEEEtran}
\bibliography{paperbib.tex}
%



%


\begin{IEEEbiography}[{\includegraphics[width=1in,height=1.25in,clip,keepaspectratio]{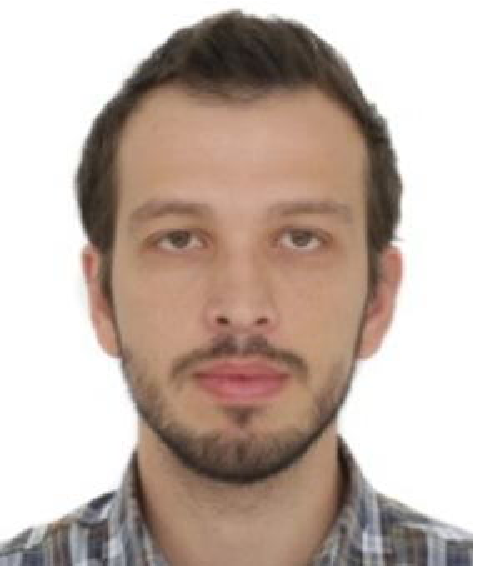}}]{Panagiotis Tzirakis}
is a first year PhD student in the High Performance Embedded and Distributed Systems (HiPEDS) Centre for Doctoral Training (CDT) programme of Imperial College London. His research interests are in the areas of deep learning, face identification, and speech recognition.
\end{IEEEbiography}
\vspace{-1cm}

\begin{IEEEbiography}[{\includegraphics[width=1in,height=1.25in,clip,keepaspectratio]{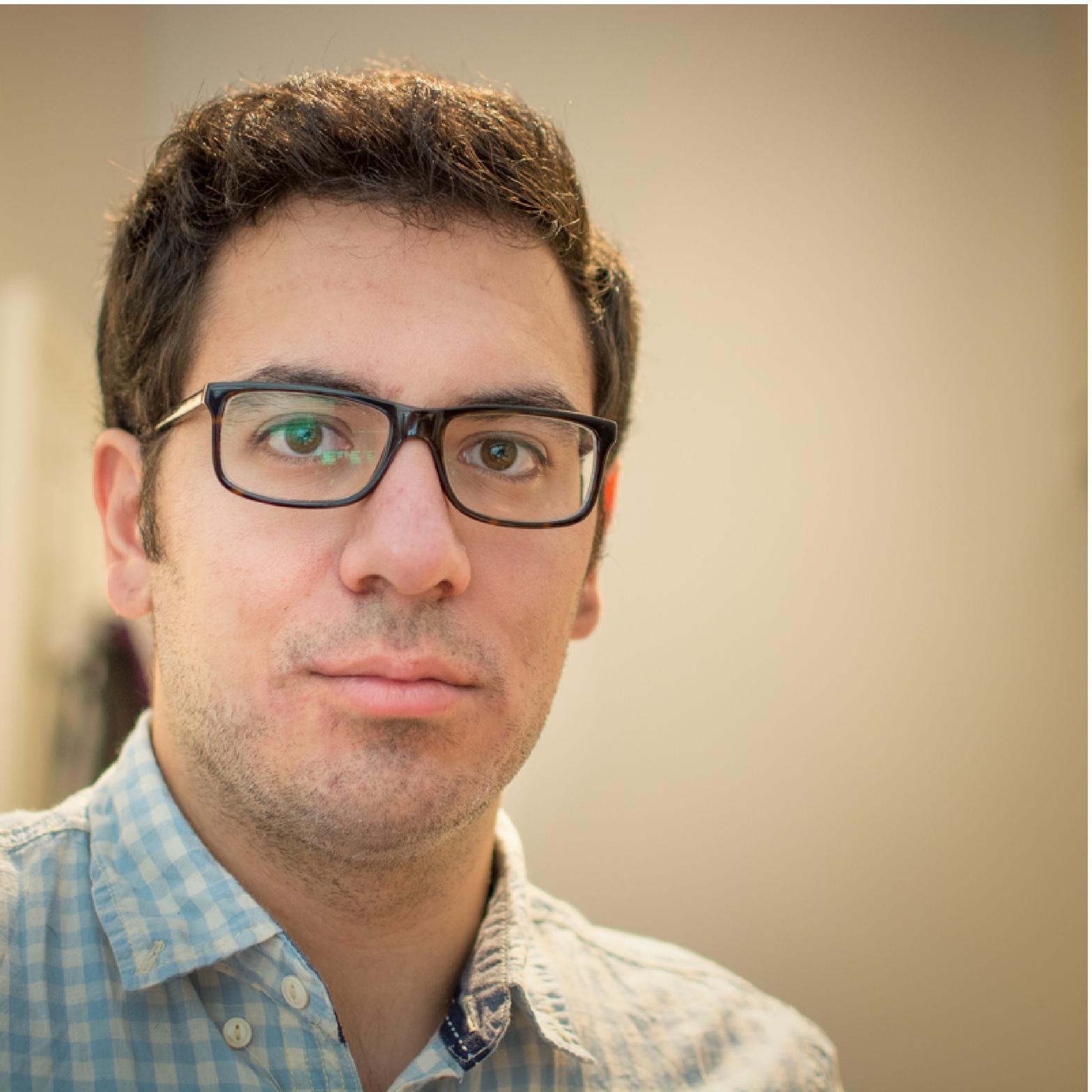}}]{George Trigeorgis}
 has received an MEng in Artificial Intelligence in 2013 from the Department of Computing, Imperial College London, where he is currently completing his Ph.D. studies.
\end{IEEEbiography}
\vspace{-1cm}

\begin{IEEEbiography}[{\includegraphics[width=1in,height=1.25in,clip,keepaspectratio]{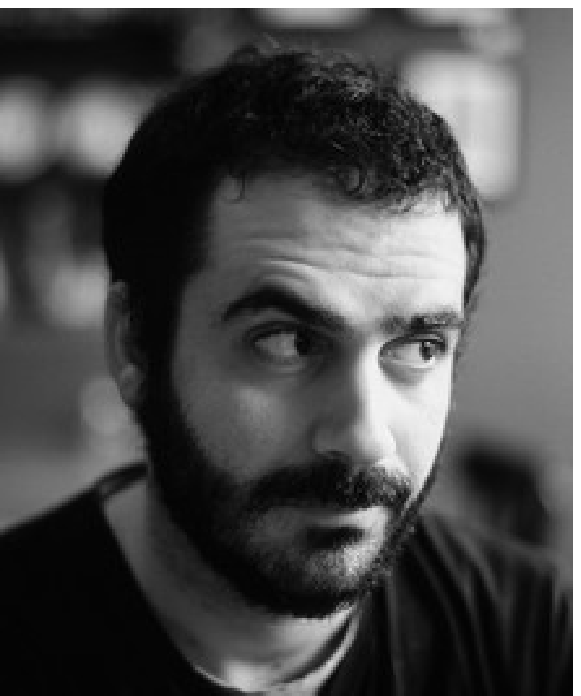}}]{Mihalis A. Nicolaou}
is a Lecturer at the Department of Computing at Goldsmiths, University of London and an Honorary Research Fellow with the Department of Computing at Imperial College London.  Mihalis obtained his PhD from the same department at Imperial, while he completed his undergraduate studies at the Department of Informatics and Telecommunications at the University of Athens, Greece.  Mihalis' research interests span the areas of machine learning, computer vision and affective computing.  He has been the recipient of several awards and scholarships for his research, including a Best Paper Award at IEEE FG, while publishing extensively in related prestigious venues.   Mihalis served as a Guest Associate Editor for the IEEE Transactions on Affective Computing and is a member of the IEEE.
\end{IEEEbiography}

\begin{IEEEbiography}[{\includegraphics[width=1in,height=1.25in,clip,keepaspectratio]{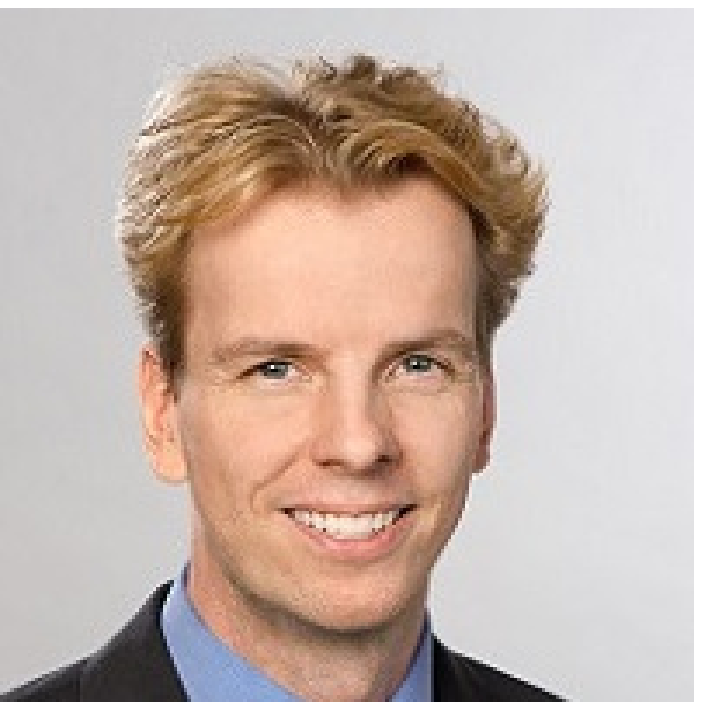}}]{Bj\"orn~W.~Schuller}
received his diploma,
doctoral degree, habilitation, and Adjunct Teaching Professorship
all in EE/IT from TUM in Munich/Germany.
At present, he is a Reader (Associate Professor) in Machine Learning
at Imperial College London/UK, Full Professor and Chair of Complex
\& Intelligent Systems at the University of Passau/Germany,
and the co-founding CEO of audEERING GmbH.
Previously, he headed the Machine Intelligence and Signal Processing Group at TUM from 2006 to
2014. In 2013 he was also invited as a permanent Visiting Professor at the Harbin Institute
of Technology/P.R. China and the University of Geneva/Switzerland.
In 2012 he was with Joanneum Research in Graz/Austria remaining an expert consultant.
In 2011 he was guest lecturer in Ancona/Italy and visiting researcher in the
Machine Learning Research Group of NICTA in Sydney/Australia. From
2009 to 2010 he was with the CNRS-LIMSI in Orsay/France, and a visiting scientist at Imperial College.
He co-authored 600+ technical contributions (14,000+ citations, h-index = 56) in the field.
\end{IEEEbiography}

\begin{IEEEbiography}[{\includegraphics[width=1in,height=1.25in,clip,keepaspectratio]{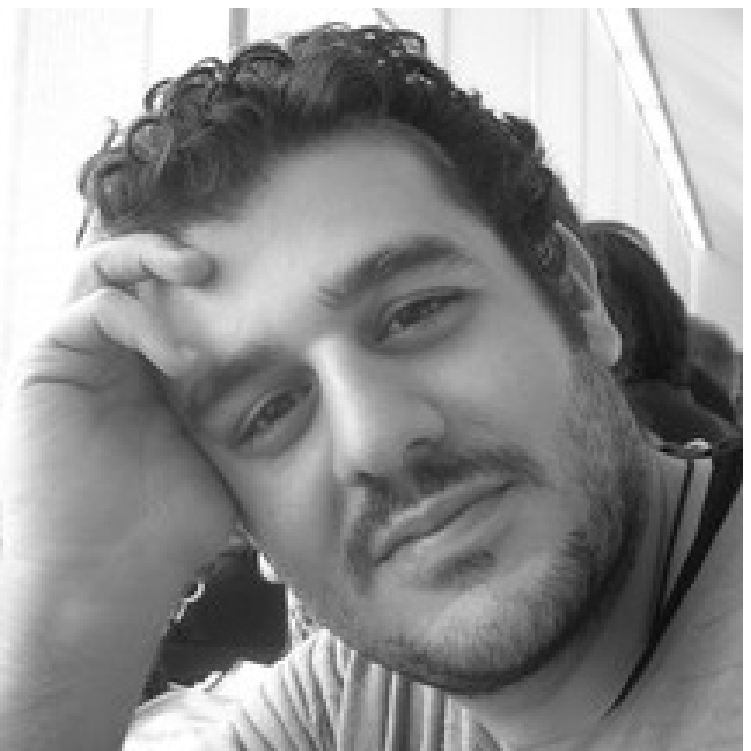}}]{Stefanos Zafeiriou}
is currently a Reader in Pattern Recognition/Statistical Machine Learning  for Computer Vision with the Department of Computing, Imperial College London, London, U.K, and a Distinguishing Research Fellow with University of Oulu under Finish Distinguishing Professor Programme. He was a recipient of the Prestigious Junior Research Fellowships from Imperial College London in 2011 to start his own independent research group. He was the recipient of the President's Medal for Excellence in Research Supervision for 2016. He has received various awards during his doctoral and post-doctoral studies. He currently serves as an Associate Editor of the IEEE Transactions on Cybernetics the Image and Vision Computing Journal. He has been a Guest Editor of over six journal special issues and co-organised over nine workshops/special sessions on face analysis topics in top venues, such as CVPR/FG/ICCV/ECCV (including two very successfully challenges run in ICCV’13 and ICCV’15 on facial landmark localisation/tracking). He has more than 2800 citations to his work, h-index 27. He is the General Chair of BMVC 2017.
\end{IEEEbiography}
\vspace{-.9cm}




\end{document}

%% file: mihalisan.tex
\usepackage{tikz}
\usetikzlibrary{arrows}
\usetikzlibrary{positioning}
\usetikzlibrary{calc}

\usepackage{pdfcomment}
\usepackage{cleveref}

\newcommand{\comments}[1]{}

\newlength\figureheight 
\newlength\figurewidth 
\newlength\mylinewidth 











%% file: bare_jrnl.bbl
\begin{thebibliography}{10}
\providecommand{\url}[1]{#1}
\csname url@samestyle\endcsname
\providecommand{\newblock}{\relax}
\providecommand{\bibinfo}[2]{#2}
\providecommand{\BIBentrySTDinterwordspacing}{\spaceskip=0pt\relax}
\providecommand{\BIBentryALTinterwordstretchfactor}{4}
\providecommand{\BIBentryALTinterwordspacing}{\spaceskip=\fontdimen2\font plus
\BIBentryALTinterwordstretchfactor\fontdimen3\font minus
  \fontdimen4\font\relax}
\providecommand{\BIBforeignlanguage}[2]{{%
\expandafter\ifx\csname l@#1\endcsname\relax
\typeout{** WARNING: IEEEtran.bst: No hyphenation pattern has been}%
\typeout{** loaded for the language `#1'. Using the pattern for}%
\typeout{** the default language instead.}%
\else
\language=\csname l@#1\endcsname
\fi
#2}}
\providecommand{\BIBdecl}{\relax}
\BIBdecl

\bibitem{ji2004real}
Q.~Ji, Z.~Zhu, and P.~Lan, ``Real-time nonintrusive monitoring and prediction
  of driver fatigue,'' \emph{IEEE Transactions on Vehicular Technology},
  vol.~53, no.~4, pp. 1052--1068, 2004.

\bibitem{burkhardt2006detecting}
F.~Burkhardt, J.~Ajmera, R.~Englert, J.~Stegmann, and W.~Burleson, ``Detecting
  anger in automated voice portal dialogs,'' in \emph{Proceedings of the Annual
  Conference of the International Speech Communication Association},
  Pittsburgh, United States, September 2006, pp. 1053--1056.

\bibitem{anagnostopoulos2015features}
C.-N. Anagnostopoulos, T.~Iliou, and I.~Giannoukos, ``Features and classifiers
  for emotion recognition from speech: a survey from 2000 to 2011,''
  \emph{Artificial Intelligence Review}, vol.~43, no.~2, pp. 155--177, 2015.

\bibitem{hinton2012deep}
G.~Hinton, L.~Deng, Y.~Dong, G.~Dahl, A.~Mohamed, N.~Jaitly, A.~Senior,
  V.~Vanhoucke, P.~Nguyen, T.~Sainath, and B.~Kingsbury, ``Deep neural networks
  for acoustic modeling in speech recognition: The shared views of four
  research groups,'' \emph{Signal Processing Magazine}, vol.~29, no.~6, pp.
  82--97, November 2012.

\bibitem{graves2014towards}
A.~Graves and N.~Jaitly, ``Towards end-to-end speech recognition with recurrent
  neural networks,'' in \emph{Proceedings of International Conference on
  Machine Learning}, Beijing, China, June 2014, pp. 1764--1772.

\bibitem{Schuller13-TI2}
B.~Schuller, S.~Steidl, A.~Batliner, A.~Vinciarelli, K.~Scherer, F.~Ringeval,
  M.~Chetouani, F.~Weninger, F.~Eyben, E.~Marchi, M.~Mortillaro, H.~Salamin,
  A.~Polychroniou, F.~Valente, and S.~Kim, ``{The INTERSPEECH 2013
  Computational Paralinguistics Challenge: Social signals, conflict, emotion,
  autism},'' in \emph{Proceedings of the Annual Conference of the International
  Speech Communication Association}, Lyon, France, August 2013, pp. 148--152.

\bibitem{avec2015}
F.~Ringeval, B.~Schuller, M.~Valstar, S.~Jaiswal, E.~Marchi, D.~Lalanne,
  R.~Cowie, and M.~Pantic, ``{AV+EC 2015 -- The First Affect Recognition
  Challenge Bridging Across Audio, Video, and Physiological Data},'' in
  \emph{Proceedings of the 5th International Workshop on Audio/Visual Emotion
  Challenge}, Brisbane, Australia, October 2015, pp. 3--8.

\bibitem{he2016deep}
K.~He, X.~Zhang, S.~Ren, and J.~Sun, ``Deep residual learning for image
  recognition,'' in \emph{Proceedings of the Conference on Computer Vision and
  Pattern Recognition}, Las Vegas, United States, June-July 2016, pp. 770--778.

\bibitem{Lin89-ACC}
L.~I.-K. Lin, ``A concordance correlation coefficient to evaluate
  reproducibility,'' \emph{Biometrics}, vol.~45, no.~1, pp. 255--268, March
  1989.

\bibitem{trigeorgis2016adieu}
G.~Trigeorgis, F.~Ringeval, R.~Brueckner, E.~Marchi, M.~A. Nicolaou,
  B.~Schuller, and S.~Zafeiriou, ``Adieu features? end-to-end speech emotion
  recognition using a deep convolutional recurrent network,'' in
  \emph{Proceedings of the IEEE International Conference on Acoustics, Speech
  and Signal Processing}, Shanghai, China, March 2016, pp. 5200--5204.

\bibitem{zemel1994autoencoders}
R.~S. Zemel, ``Autoencoders, minimum description length and helmholtz free
  energy,'' in \emph{Proceedings of the Neural Information Processing Systems},
  1994.

\bibitem{lecun1989generalization}
Y.~LeCun \emph{et~al.}, ``Generalization and network design strategies,''
  \emph{Connectionism in Perspective}, pp. 143--155, 1989.

\bibitem{hinton2006fast}
G.~E. Hinton, S.~Osindero, and Y.-W. Teh, ``A fast learning algorithm for deep
  belief nets,'' \emph{Neural Computation}, vol.~18, no.~7, pp. 1527--1554,
  2006.

\bibitem{hochreiter1997long}
S.~Hochreiter and J.~Schmidhuber, ``Long short-term memory,'' \emph{Neural
  Computation}, vol.~9, no.~8, pp. 1735--1780, 1997.

\bibitem{ngiam2011multimodal}
J.~Ngiam, A.~Khosla, M.~Kim, J.~Nam, H.~Lee, and A.~Y. Ng, ``Multimodal deep
  learning,'' in \emph{Proceedings of the International Conference on Machine
  Learning}, Washington, United States, June-July 2011, pp. 689--696.

\bibitem{hu2016temporal}
D.~Hu, X.~Li \emph{et~al.}, ``Temporal multimodal learning in audiovisual
  speech recognition,'' in \emph{Proceedings of the IEEE Conference on Computer
  Vision and Pattern Recognition}, Las Vegas, United States, June-July 2016,
  pp. 3574--3582.

\bibitem{wu2016deep}
D.~Wu, L.~Pigou, P.-J. Kindermans, N.~D.-H. Le, L.~Shao, J.~Dambre, and J.-M.
  Odobez, ``Deep dynamic neural networks for multimodal gesture segmentation
  and recognition,'' \emph{IEEE Transactions on Pattern Analysis and Machine
  Intelligence}, vol.~38, no.~8, pp. 1583--1597, 2016.

\bibitem{han2014speech}
K.~Han, D.~Yu, and I.~Tashev, ``Speech emotion recognition using deep neural
  network and extreme learning machine.'' in \emph{Proceedings of the Annual
  Conference of the International Speech Communication Association}, Singapore,
  September 2014, pp. 223--227.

\bibitem{lim2016speech}
W.~Lim, D.~Jang, and T.~Lee, ``Speech emotion recognition using convolutional
  and recurrent neural networks,'' in \emph{Proceedings of the Signal and
  Information Processing Association Annual Summit and Conference}, Jeju,
  Korea, December 2016, pp. 1--4.

\bibitem{huang2016deep}
Y.~Huang and H.~Lu, ``Deep learning driven hypergraph representation for
  image-based emotion recognition,'' in \emph{Proceedings of the International
  Conference on Multimodal Interaction}, Tokyo, Japan, November 2016, pp.
  243--247.

\bibitem{ebrahimi2015recurrent}
S.~Ebrahimi~Kahou, V.~Michalski, K.~Konda, R.~Memisevic, and C.~Pal,
  ``Recurrent neural networks for emotion recognition in video,'' in
  \emph{Proceedings of the International Conference on Multimodal Interaction},
  Seattle, United States, November 2015, pp. 467--474.

\bibitem{liu2016emotion}
W.~Liu, W.-L. Zheng, and B.-L. Lu, ``Emotion recognition using multimodal deep
  learning,'' in \emph{International Conference on Neural Information
  Processing}, Kyoto, Japan, October 2016, pp. 521--529.

\bibitem{kim2013deep}
Y.~Kim, H.~Lee, and E.~M. Provost, ``Deep learning for robust feature
  generation in audiovisual emotion recognition,'' in \emph{Proceedings of the
  International Conference on Acoustics, Speech and Signal Processing},
  Vancouver, Canada, May 2013, pp. 3687--3691.

\bibitem{kahou2016emonets}
S.~E. Kahou, X.~Bouthillier, P.~Lamblin, C.~Gulcehre, V.~Michalski, K.~Konda,
  S.~Jean, P.~Froumenty, Y.~Dauphin, N.~Boulanger-Lewandowski \emph{et~al.},
  ``Emonets: Multimodal deep learning approaches for emotion recognition in
  video,'' \emph{Journal on Multimodal User Interfaces}, vol.~10, no.~2, pp.
  99--111, 2016.

\bibitem{sun2016exploring}
B.~Sun, S.~Cao, L.~Li, J.~He, and L.~Yu, ``Exploring multimodal visual features
  for continuous affect recognition,'' in \emph{Proceedings of the 6th
  International Workshop on Audio/Visual Emotion Challenge}, Amsterdam, The
  Netherlands, October 2016, pp. 83--88.

\bibitem{Zhang}
S.~Zhang, S.~Zhang, T.~Huang, and W.~Gao, ``Multimodal deep convolutional
  neural network for audio-visual emotion recognition,'' in \emph{Proceedings
  of the International Conference on Multimedia Retrieval}, New York, United
  States, June 2016, pp. 281--284.

\bibitem{ringeval2015prediction}
F.~Ringeval, F.~Eyben, E.~Kroupi, A.~Yuce, J.-P. Thiran, T.~Ebrahimi,
  D.~Lalanne, and B.~Schuller, ``Prediction of asynchronous dimensional emotion
  ratings from audiovisual and physiological data,'' \emph{Pattern Recognition
  Letters}, vol.~66, pp. 22--30, 2015.

\bibitem{han2016}
J.~Han, Z.~Zhang, N.~Cummins, F.~Ringeval, and B.~Schuller, ``Strength
  modelling for real-worldautomatic continuous affect recognition from
  audiovisual signals,'' \emph{Image and Vision Computing}, pp.~--, 2016.

\bibitem{valstar2016avec}
M.~Valstar, J.~Gratch, B.~Schuller, F.~Ringeval, D.~Lalanne, M.~Torres~Torres,
  S.~Scherer, G.~Stratou, R.~Cowie, and M.~Pantic, ``Avec 2016: Depression,
  mood, and emotion recognition workshop and challenge,'' in \emph{Proceedings
  of the 6th International Workshop on Audio/Visual Emotion Challenge},
  Amsterdam, The Netherlands, October 2016, pp. 3--10.

\bibitem{huang2016staircase}
Z.~Huang, B.~Stasak, T.~Dang, K.~Wataraka~Gamage, P.~Le, V.~Sethu, and J.~Epps,
  ``Staircase regression in oa rvm, data selection and gender dependency in
  avec 2016,'' in \emph{Proceedings of the 6th International Workshop on
  Audio/Visual Emotion Challenge}, Amsterdam, The Netherlands, October 2016,
  pp. 19--26.

\bibitem{Weber2016}
R.~Weber, V.~Barrielle, C.~Soladi{\'e}, and R.~S{\'e}guier, ``High-level
  geometry-based features of video modality for emotion prediction,'' in
  \emph{Proceedings of the 6th International Workshop on Audio/Visual Emotion
  Challenge}, Amsterdam, The Netherlands, October 2016, pp. 51--58.

\bibitem{brady2016multi}
K.~Brady, Y.~Gwon, P.~Khorrami, E.~Godoy, W.~Campbell, C.~Dagli, and T.~S.
  Huang, ``Multi-modal audio, video and physiological sensor learning for
  continuous emotion prediction,'' in \emph{Proceedings of the 6th
  International Workshop on Audio/Visual Emotion Challenge}, Amsterdam, The
  Netherlands, October 2016, pp. 97--104.

\bibitem{povolny2016multimodal}
F.~Povolny, P.~Matejka, M.~Hradis, A.~Popkov{\'a}, L.~Otrusina, P.~Smrz,
  I.~Wood, C.~Robin, and L.~Lamel, ``Multimodal emotion recognition for avec
  2016 challenge,'' in \emph{Proceedings of the 6th International Workshop on
  Audio/Visual Emotion Challenge}, Amsterdam, The Netherlands, October 2016,
  pp. 75--82.

\bibitem{somandepalli2016online}
K.~Somandepalli, R.~Gupta, M.~Nasir, B.~M. Booth, S.~Lee, and S.~S. Narayanan,
  ``Online affect tracking with multimodal kalman filters,'' in
  \emph{Proceedings of the International Workshop on Audio/Visual Emotion
  Challenge}, 2016, pp. 59--66.

\bibitem{hirsch1991improved}
H.~Hirsch, P.~Meyer, and H.~Ruehl, ``Improved speech recognition using
  high-pass filtering of subband envelopes,'' in \emph{Proceedings of the
  European Conference on Speech Technology}, Genoa, Italy, September 1991, pp.
  413--416.

\bibitem{schluter2007gammatone}
R.~Schl{\"u}ter, L.~Bezrukov, H.~Wagner, and H.~Ney, ``Gammatone features and
  feature combination for large vocabulary speech recognition,'' in
  \emph{Proceedings of the IEEE International Conference on Acoustics, Speech
  and Signal Processing}, Honolulu, United States, April 2007, pp. 649--652.

\bibitem{Ringeval15-POA}
F.~Ringeval, F.~Eyben, E.~Kroupi, A.~Yuce, J.-P. Thiran, T.~Ebrahimi,
  D.~Lalanne, and B.~Schuller, ``Prediction of asynchronous dimensional emotion
  ratings from audiovisual and physiological data,'' \emph{Pattern Recognition
  Letters}, vol.~66, pp. 22--30, November 2015.

\bibitem{russakovsky2015imagenet}
O.~Russakovsky, J.~Deng, H.~Su, J.~Krause, S.~Satheesh, S.~Ma, Z.~Huang,
  A.~Karpathy, A.~Khosla, M.~Bernstein \emph{et~al.}, ``Imagenet large scale
  visual recognition challenge,'' \emph{International Journal of Computer
  Vision}, vol. 115, no.~3, pp. 211--252, 2015.

\bibitem{Ringeval13-ITR}
S.-A. S.~J. Ringeval, Fabien and D.~Lalanne, ``{Introducing the RECOLA
  Multimodal Corpus of Remote Collaborative and Affective Interactions},'' in
  \emph{Proceedings of the IEEE International Conference and Workshops on
  Automatic Face and Gesture Recognition}, Shanghai, China, April 2013, pp.
  1--8.

\bibitem{kingma2014adam}
D.~Kingma and J.~Ba, ``Adam: A method for stochastic optimization,''
  \emph{arXiv preprint arXiv:1412.6980}, 2014.

\bibitem{srivastava2014dropout}
N.~Srivastava, G.~Hinton, A.~Krizhevsky, I.~Sutskever, and R.~Salakhutdinov,
  ``Dropout: A simple way to prevent neural networks from overfitting,''
  \emph{Journal of Machine Learning Research}, vol.~15, no.~1, pp. 1929--1958,
  January 2014.

\bibitem{Kachele15-EMF}
M.~K\"achele, P.~Thiam, G.~Palm, F.~Schwenker, and M.~Schels, ``Ensemble
  methods for continuous affect recognition: Multimodality, temporality, and
  challenges,'' in \emph{Proceedings of the 5th International Workshop on
  Audio/Visual Emotion Challenge}, Brisbane, Australia, October 2015, pp.
  9--16.

\bibitem{Mariooryad14-CTC}
S.~Mariooryad and C.~Busso, ``Correcting time-continuous emotional labels by
  modeling the reaction lag of evaluators,'' \emph{IEEE Transactions on
  Affective Computing}, vol.~6, no.~2, pp. 97--108, April-June 2015.

\bibitem{cho2014learning}
K.~Cho, B.~Van~Merri{\"e}nboer, C.~Gulcehre, D.~Bahdanau, F.~Bougares,
  H.~Schwenk, and Y.~Bengio, ``Learning phrase representations using rnn
  encoder-decoder for statistical machine translation,'' \emph{arXiv preprint
  arXiv:1406.1078}, 2014.

\bibitem{Scherer03-VCO}
K.~Scherer, ``Vocal communication of emotion: A review of research paradigms,''
  \emph{Speech Communication}, vol.~40, no. 1-2, pp. 227--256, April 2003.

\bibitem{Eyben15-TGM}
F.~Eyben, K.~R. Scherer, B.~W. Schuller, J.~Sundberg, E.~Andr{\'e}, C.~Busso,
  L.~Y. Devillers, J.~Epps, P.~Laukka, S.~S. Narayanan \emph{et~al.}, ``The
  geneva minimalistic acoustic parameter set (gemaps) for voice research and
  affective computing,'' \emph{IEEE Transactions on Affective Computing},
  vol.~7, no.~2, pp. 190--202, 2016.

\bibitem{nam2016mdnet}
H.~Nam and B.~Han, ``Learning multi-domain convolutional neural networks for
  visual tracking,'' in \emph{Proceedings of the IEEE Conference on Computer
  Vision and Pattern Recognition}, Las Vegas, United States, June 2016, pp.
  4293--4302.

\end{thebibliography}
